\documentclass[twoside,11pt]{article}

\usepackage[abbrvbib, preprint]{jmlr2e}

\usepackage{lastpage}
\usepackage{booktabs}
\usepackage{pifont}
\usepackage{adjustbox}
\usepackage{mathtools}

\usepackage[acronym, toc, nonumberlist]{glossaries}
\glsdisablehyper
\usepackage{tablefootnote}
\usepackage{xcolor}
\usepackage{subcaption}
\usepackage{placeins}

\hypersetup{hidelinks}
\jmlrheading{27}{2026}{1-\pageref{LastPage}}{10/25; Revised 7/26}{7/26}{25-2467}{Louis Berthier, Ahmed Shokry, Maxime Moreaud, Guillaume Ramelet, and Eric Moulines}
\ShortHeadings{torchsom: The Reference PyTorch Library for Self-Organizing Maps}{Berthier, Shokry, Moreaud, Ramelet, and Moulines}
\firstpageno{1}

\definecolor{ForestGreen}{RGB}{34,139,34}
\definecolor{BrickRed}{RGB}{203,65,84}
\newcommand{\cmark}{\textcolor{ForestGreen}{\ding{51}}}
\newcommand{\xmark}{\textcolor{BrickRed}{\ding{55}}}
\newcommand{\best}[1]{{\boldmath\textbf{#1}}}

\graphicspath{{assets/}}

\usepackage{tikz}
\usetikzlibrary{shapes.geometric}
\usetikzlibrary{decorations.pathreplacing}

\newcommand{\fillrect}[3]{%
    \fill[#3] (#1+0.01,#2+0.01) rectangle ++ (0.98,0.98);
}

\colorlet{bmucolor}{black!80}
\colorlet{order1color}{blue!60}
\colorlet{order2color}{teal!60}
\colorlet{order3color}{purple!40}
 
\newacronym{som}{SOM}{Self-Organizing Map}
\newacronym{jitl}{JITL}{Just-In-Time Learning}
\newacronym{ml}{ML}{Machine Learning}
\newacronym{dl}{DL}{Deep Learning}
\newacronym{bmu}{BMU}{Best Matching Unit}
\newacronym{qe}{QE}{Quantization Error}
\newacronym{te}{TE}{Topographic Error}
 
\begin{document}

\title{\texttt{torchsom}: The Reference PyTorch Library for Self-Organizing Maps}
\author{\name Louis Berthier$^{1,2}$
  \email louis-desire-romeo.berthier@michelin.com
  \AND \name Ahmed Shokry$^{1}$
  \email ahmed.shokry@polytechnique.edu
  \AND \name Maxime Moreaud$^{2}$
  \email maxime.moreaud@michelin.com
  \AND \name Guillaume Ramelet$^{2}$
  \email guillaume.ramelet@michelin.com
  \AND \name Eric Moulines$^{1}$
  \email eric.moulines@polytechnique.edu
  \AND \addr $^1$Centre de Mathématiques Appliquées, Ecole Polytechnique, 91120 Palaiseau, France
  \AND \addr $^2$Manufacture Française des Pneumatiques Michelin, 63100 Clermont-Ferrand, France
}
\editor{Alexandre Gramfort}

\maketitle

\begin{abstract}%
  This paper introduces \texttt{torchsom}, an open-source Python library that provides a reference implementation of the \gls{som} in \texttt{PyTorch}.
  This package offers three main features:
  (i) dimensionality reduction,
  (ii) clustering, and
  (iii) friendly data visualization.
  It relies on a \texttt{PyTorch} backend, enabling
  (i) fast and efficient training of \glspl{som} through GPU acceleration, and
  (ii) easy and scalable integration with the \texttt{PyTorch} ecosystem.
  \texttt{torchsom} also follows the \texttt{scikit-learn} API for ease of use and extensibility.
  The library is released under the \texttt{Apache 2.0} license with 90\% test coverage, and its source code and documentation are available at \url{https://github.com/michelin/TorchSOM}.
\end{abstract}%

\begin{keywords}
  self-organizing maps,
  pytorch,
  unsupervised learning,
  dimensionality reduction,
  clustering
\end{keywords}

\section{Introduction} \label{sec:introduction}

\glsdisp{som}{Self-Organizing Maps (SOMs)} remain a valuable and enduring technique in modern \glsdisp{ml}{machine learning (ML)} and data analytics, despite being introduced decades ago \citep{kohonenSelforganizedFormationTopologically1982a,kohonenSelforganizingMap1990,kohonenSelfOrganizingMaps2001a}.
This is because \glspl{som} integrate key mechanisms at low computational cost, including:
(i) abstraction of high-dimensional data with dimensionality reduction,
(ii) preservation of latent nonlinear topological structures, and
(iii) visual interpretability.
Such properties make \glspl{som} a valuable asset for exploratory analysis, explainable AI workflows, and resource-constrained environments.

These properties have driven the extensive adoption of \glspl{som} in a broad range of domains, such as
energy industry \citep{rajKeyGasesTransformer2023b, concettiUnsupervisedAnomalyDetection2023, dashPerformanceAssessmentDifferent2024},
biology and health \citep{haoSOMDEScalableMethod2021a, farzamniaMRIBrainTumor2023, weberApplicationSelforganizingMaps2023},
IoT systems \citep{khanDiscoverBotnetsIoT2023, gadJointSelfOrganizingMaps2024},
chemical and environmental applications \citep{maltarolloApplicationsArtificialNeural2013, fengAnalysisWaterQuality2023, xiangPotentialEcologicalRisk2022, miaAnalysisSelforganizingMaps2023, zhangHydrogeochemicalAnalysisGroundwater2023, licenSelforganizingMapAlgorithm2023},
and business cases \citep{bloomMARKETSEGMENTATIONNeural2005, bowenSelforganizingMapsNovel2024}.
\glspl{som} can also be periodically updated with new data, progressively improving their representations and supporting applications such as industrial monitoring and control \citep{zhangOnlineSemisupervisedQuality2018,zhengJustintimeSemisupervisedSoft2018, jinEnsembleJustInTimeLearningBased2020, urhanIntegratingAdaptiveMoving2020,zhangDeepSemiSupervisedJustinTime2022}.

Despite this relevance, only a few Python libraries attempt to bridge \glspl{som} with modern \gls{ml} workflows, such as integration with \texttt{PyTorch} \citep{anselPyTorch2Faster2024} or a \texttt{scikit-learn} interface \citep{pedregosaScikitlearnMachineLearning2011}.
However, these limited implementations are often outdated and poorly maintained, and they lack
(i) GPU acceleration,
(ii) integration with modern \gls{dl} frameworks, and
(iii) user-friendly APIs with visualization capabilities.
Consequently, the Python \gls{som} ecosystem still suffers from significant gaps, hindering reproducible and scalable \gls{som}-based analyses.

To overcome these challenges, we introduce \texttt{torchsom}, to the best of our knowledge the most complete library for \glspl{som} built on the \texttt{PyTorch} ecosystem.
\texttt{torchsom} integrates GPU acceleration, \gls{som} variants, clustering tools, and a familiar \texttt{scikit-learn}-style API, complemented by user-friendly visualization tools.
Our library is designed to integrate into \gls{ml} workflows, combining modern \gls{dl} frameworks with sound software engineering practices, while improving efficiency and strengthening the visual interpretability that distinguishes \glspl{som}, whose underlying principles are summarized in Appendix~\ref{app:som_overview}. 

\section{Related Work} \label{sec:related_work}

In the Python ecosystem, several libraries provide \gls{som} implementations.
However, they differ in their technical architecture, development, and capabilities as shown in Table~\ref{tab:som_libraries}.

\begin{table}[tbp]
    \centering
    \begin{adjustbox}{width=\textwidth}
        \begin{tabular}{@{}lcccccc@{}}
            \toprule
                                        & \shortstack{\texttt{torchsom}                                                                                                                                                                                                                                                                                                                                                                                                                                \\ \citep{berthierTorchSOMReferencePyTorch2025}} & \shortstack{\texttt{MiniSom} \\ \citep{vettigliMiniSomMinimalisticNumPybased2018}} & \shortstack{\texttt{SimpSOM} \\ \citep{comitaniSimpSOMSimpleSelfOrganizing2017a}} & \shortstack{\texttt{SOMPY} \\ \citep{moosaviPythonLibrarySelf2014}} & \shortstack{\texttt{somoclu} \\ \citep{wittekSomocluMassivelyParallel2017}} & \shortstack{\texttt{som-pbc} \\ \citep{mullerSompbcSimpleSelforganizing2018}} \\
            \midrule
            \multicolumn{7}{l}{\textit{Technical Architecture}}                                                                                                                                                                                                                                                                                                                                                                                                                                        \\
            \textbf{Framework}          & PyTorch                       & NumPy                                                                                                                                                                                                                                                                  & NumPy     & NumPy    & C++                                                                                                                 & NumPy  \\
            \textbf{GPU Acceleration}   & CUDA PyTorch                  & \xmark                                                                                                                                                                                                                                                                 & CuPy/cuML & \xmark   & CUDA C++                                                                                                            & \xmark \\
            \textbf{JIT Compilation}    & via PyTorch                   & Numba\tablefootnote{Opt-in Numba~\citep{lamNumbaLLVMbasedPython2015} JIT (\texttt{train\_batch\_offline\_fast}); distinct from \emph{Just-In-Time Learning} (\gls{jitl}, below), which is a supervised local-modeling capability rather than a compilation technique.} & \xmark    & \xmark   & \xmark~(AOT)\tablefootnote{\texttt{somoclu}'s kernels are ahead-of-time compiled with \texttt{nvcc} at build time.} & \xmark \\
            \textbf{API Design}         & scikit-learn                  & Custom                                                                                                                                                                                                                                                                 & Custom    & MATLAB   & Custom                                                                                                              & Custom \\
            \midrule
            \multicolumn{7}{l}{\textit{Development}}                                                                                                                                                                                                                                                                                                                                                                                                                                                   \\
            \textbf{Maintenance}        & Active                        & Active                                                                                                                                                                                                                                                                 & Minimal   & Minimal  & Minimal                                                                                                             & \xmark \\
            \textbf{Documentation}      & Rich                          & Basic\tablefootnote{Example notebooks and partial in-code docstrings; no narrative documentation site (e.g. comparable to \url{https://opensource.michelin.io/TorchSOM}).}                                                                                             & Basic     & \xmark   & Basic                                                                                                               & Basic  \\
            \textbf{Test Coverage}      & $90\%$                        & $98\%$                                                                                                                                                                                                                                                                 & $53\%$    & \xmark   & Minimal                                                                                                             & \xmark \\
            \textbf{PyPI Distribution}  & \cmark                        & \cmark                                                                                                                                                                                                                                                                 & \cmark    & \xmark   & \cmark                                                                                                              & \xmark \\
            \midrule
            \multicolumn{7}{l}{\textit{Functional Capabilities (built-in)}}                                                                                                                                                                                                                                                                                                                                                                                                                            \\
            \textbf{Visualization}      & Advanced                      & \xmark                                                                                                                                                                                                                                                                 & Moderate  & Moderate & Basic                                                                                                               & Basic  \\
            \textbf{Clustering}         & \cmark                        & Examples only\tablefootnote{Clustering is not a built-in \texttt{MiniSom} feature; it requires user-supplied code on top of \texttt{MiniSom} primitives.}                                                                                                              & \cmark    & \xmark   & \xmark                                                                                                              & \xmark \\
            \textbf{\gls{jitl} support} & \cmark                        & \xmark                                                                                                                                                                                                                                                                 & \xmark    & \xmark   & \xmark                                                                                                              & \xmark \\
            \textbf{SOM Variants}       & Multiple                      & \xmark                                                                                                                                                                                                                                                                 & PBC       & \xmark   & PBC                                                                                                                 & PBC    \\
            \textbf{Extensibility}      & High                          & Moderate                                                                                                                                                                                                                                                               & Low       & Low      & Low                                                                                                                 & Low    \\
            \bottomrule
        \end{tabular}
    \end{adjustbox}
    \caption{Comparative analysis of Python SOM libraries available on GitHub.}
    \label{tab:som_libraries}
\end{table}

While existing libraries address specific use cases (\texttt{MiniSom} offers a minimalist \texttt{NumPy}-based implementation suited to education and prototyping, and \texttt{somoclu} targets HPC environments through CUDA C++), \texttt{torchsom} is the only library in this comparison that combines a native \texttt{PyTorch} backend with GPU acceleration, a \texttt{scikit-learn}-compatible API, an advanced built-in visualization suite, a built-in clustering interface, \gls{jitl} support, and multiple grid topologies with configurable neighborhood retrieval modes within a single modular codebase.
It is further supported by a published narrative documentation site (\url{https://opensource.michelin.io/TorchSOM}) and a community-oriented development process, making \texttt{torchsom} a complete and scalable reference for both research and production.

\section{Package Architecture} \label{sec:architecture}

The \texttt{torchsom} library follows a modular design built around three core components that provide a complete \gls{som} implementation with native \texttt{PyTorch} integration, namely:

\begin{enumerate}
      \item \texttt{torchsom.core}: \\
            This module implements classical \gls{som} algorithms \citep{kohonenSelforganizingMap1990} in the \texttt{PyTorch} ecosystem for integration into \gls{dl} workflows.
            The core classes provide
            (i) a \texttt{fit()} supporting automatic GPU acceleration for model training,
            (ii) a \texttt{cluster()} for partitioning the trained map,
            (iii) a \texttt{build\_map()} for generating maps suitable for visualization, and
            (iv) a \texttt{collect\_samples()} for identifying informative samples from grid topology and latent-space distances.
            The last of these offers three configurable retrieval modes: \texttt{bmu\_only}, \texttt{bmu\_neighborhood}, and \texttt{bmu\_neighborhood\_knn}.
            \gls{bmu} selection is delegated to a configurable backend (the \texttt{search\_backend} argument), which by default selects automatically between a \texttt{PyTorch} brute-force implementation and an optional \texttt{FAISS} backend that accelerates nearest-neighbor search for large maps and high-dimensional inputs.

      \item \texttt{torchsom.utils}: \\
            This module provides essential components for \gls{som} parameterization and training, including decay functions for learning rate and neighborhood width scheduling.
            It supports multiple distance metrics, with Euclidean and cosine distances commonly used for latent space calculations and \gls{bmu} selection.
            Multiple neighborhood kernels are implemented, with the Gaussian kernel serving as the primary example for weight updates around the \gls{bmu}.
            Both rectangular and hexagonal grid topologies are supported, each optionally combined with periodic boundary conditions (the \texttt{pbc} flag) that wrap the lattice into a toroidal structure to eliminate edge effects.
            Finally, three clustering methods are available:
            (i) \texttt{K-means} \citep{ikotunKmeansClusteringAlgorithms2023},
            (ii) \texttt{GMM} \citep{liMixtureDensityEstimation1999,figueiredoUnsupervisedLearningFinite2002}, and
            (iii) \texttt{HDBSCAN} \citep{campelloDensityBasedClusteringBased2013,mcinnesHdbscanHierarchicalDensity2017}.

      \item \texttt{torchsom.visualization}: \\
            This module fills a significant gap in existing \gls{som} implementations by providing visualizations for both rectangular and hexagonal topologies.
            It includes seven visualization types:
            (i) U-matrix for map topology and cluster structure,
            (ii) hit maps showing neuron activation patterns,
            (iii) component planes for feature-wise analysis,
            (iv) classification and metric maps for target statistics,
            (v) score and rank maps for quality assessment,
            (vi) training curves to monitor convergence, and
            (vii) cluster maps with associated quality metrics.
            All visualizations are \texttt{matplotlib}-based, supporting customizable styling and automatic figure generation (see Appendix~\ref{app:visualization}).
\end{enumerate}

\texttt{torchsom} is designed to integrate with modern \gls{dl} workflows using the \texttt{PyTorch} ecosystem, enabling GPU acceleration and efficient batch processing of large data sets.
Comprehensive documentation provides implementation details, API references, and user guides covering regression, classification, and clustering tasks.\footnote{The documentation is accessible at \url{https://opensource.michelin.io/TorchSOM}.}

The modular and extensible architecture of \texttt{torchsom} allows users to add new visualizations, customize functionalities, and extend core components.
This flexibility facilitates rapid experimentation and adaptation, while also encouraging contributions from the open-source community.
By supporting integration and customization, \texttt{torchsom} serves as a reference implementation for both research and production environments, enabling users to extend and tailor \gls{som} applications to their specific requirements. 

\section{Benchmarks} \label{sec:benchmarks}

\texttt{torchsom}'s computational performance and fidelity are evaluated against \texttt{MiniSom}, the most widely adopted and actively maintained \gls{som} library, and against \texttt{somoclu}, a massively parallel C++/CUDA implementation (see Table~\ref{tab:som_libraries}).
To keep every comparison fair, backends are grouped by execution device: CPU backends are compared with one another in Table~\ref{tab:benchmarks_cpu}, and GPU backends in Table~\ref{tab:benchmarks_gpu}.

Synthetic data sets are generated using \texttt{scikit-learn}'s \texttt{make\_blobs()}, varying both sample size and feature dimensionality to assess scaling behavior.
We consider data sets with sample sizes of \{240, 4000, 16000\} and feature counts of \{4, 50, 300\}.
All implementations use identical, commonly adopted hyperparameters to ensure a fair comparison: a $25\times15$ grid, PCA initialization, rectangular topology, 100 training iterations, Gaussian neighborhood function, and Euclidean distance.
All scripts, configuration files, and seed-controlled run wrappers used to produce Tables~\ref{tab:benchmarks_cpu} and~\ref{tab:benchmarks_gpu} are publicly released under the \texttt{benchmark/} directory of the repository; the exact state of this revision is reproducible from the Git tag \texttt{jmlr-revision-v2}.

Both tables report
(i) test \gls{qe},
(ii) test \gls{te}, and
(iii) wall-clock time (initialization + training + compilation/device setup) averaged over 10 runs.
For readability, standard deviations are shown only when they are non-zero.
This convention applies to every benchmark table, here and in Appendix~\ref{app:benchmark_results}.
The CPU comparison (Table~\ref{tab:benchmarks_cpu}) considers
(a) \texttt{MiniSom} with its standard online training,
(b) \texttt{MiniSom}-JIT, its optional Numba-accelerated batch-offline routine, and
(c) \texttt{torchsom}.
The GPU comparison (Table~\ref{tab:benchmarks_gpu}) considers the two backends with a native GPU kernel:
(a) \texttt{somoclu} (CUDA C++) and
(b) \texttt{torchsom} (CUDA PyTorch).
Extensive benchmarking results, together with the metric definitions and the per-backend measurement details, are provided in Appendix~\ref{app:benchmark_results}.\footnote{%
    All benchmarks were run on an Intel Xeon Gold 6134 CPU (16 cores at 3.20 GHz, 187 GB RAM) and an NVIDIA Tesla V100-PCIE GPU (5120 CUDA cores at 1.38 GHz, 32 GB RAM).
}

\begin{table}[tb]
    \centering
    \scriptsize
    \adjustbox{max width=\linewidth}{
        \begin{tabular}{@{}rr|ccc|ccc|ccc@{}}
            \toprule
            \multicolumn{2}{c|}{Data set} & \multicolumn{3}{c|}{\texttt{MiniSom} (CPU)} & \multicolumn{3}{c|}{\texttt{MiniSom}-JIT (CPU)} & \multicolumn{3}{c}{\texttt{torchsom} (CPU)}                                                                                                                                                                                     \\
            \cmidrule(lr){1-2} \cmidrule(lr){3-5} \cmidrule(lr){6-8} \cmidrule(lr){9-11}
            Samples                       & Features                                    & \gls{qe} $\downarrow$                           & \gls{te} (\%) $\downarrow$                  & Time (s) $\downarrow$ & \gls{qe} $\downarrow$ & \gls{te} (\%) $\downarrow$ & Time (s) $\downarrow$ & \gls{qe} $\downarrow$ & \gls{te} (\%) $\downarrow$ & Time (s) $\downarrow$   \\
            \midrule
            240                           & 4                                           & \best{$0.17$}                                   & $26 \pm 4$                                  & $1.45 \pm 0.03$       & $0.18$                & $23$                       & $4.31 \pm 0.01$       & $0.24$                & \best{$2$}                 & \best{$0.27 \pm 0.04$}  \\
            240                           & 50                                          & $1.80$                                          & $48 \pm 7$                                  & $2.74 \pm 0.05$       & \best{$1.78$}         & $40$                       & $2.38 \pm 0.01$       & $1.79$                & \best{$5$}                 & \best{$0.34 \pm 0.07$}  \\
            240                           & 300                                         & $5.44$                                          & $71 \pm 10$                                 & $13.84 \pm 0.19$      & $5.34$                & $52$                       & $5.87 \pm 0.01$       & \best{$5.21$}         & \best{$27$}                & \best{$0.52 \pm 0.04$}  \\
            4000                          & 4                                           & \best{$0.16$}                                   & $32 \pm 1$                                  & $25.08 \pm 0.35$      & \best{$0.16$}         & $20$                       & $5.91 \pm 0.03$       & $0.23$                & \best{$0$}                 & \best{$3.26 \pm 0.05$}  \\
            4000                          & 50                                          & $1.66$                                          & $55 \pm 2$                                  & $49.03 \pm 1.10$      & \best{$1.64$}         & $28$                       & $16.33 \pm 0.58$      & $1.77$                & \best{$4$}                 & \best{$3.50 \pm 0.09$}  \\
            4000                          & 300                                         & $5.14$                                          & $74 \pm 2$                                  & $226 \pm 2$           & \best{$5.02$}         & $37$                       & $69.47 \pm 0.35$      & $5.13$                & \best{$6$}                 & \best{$5.25 \pm 0.10$}  \\
            16000                         & 4                                           & \best{$0.15$}                                   & $32 \pm 1$                                  & $99.35 \pm 1.53$      & \best{$0.15$}         & $19$                       & $17.58 \pm 0.05$      & $0.23$                & \best{$1$}                 & \best{$14.55 \pm 0.20$} \\
            16000                         & 50                                          & $1.64$                                          & $56 \pm 1$                                  & $186 \pm 2$           & \best{$1.60$}         & $25$                       & $51.35 \pm 0.20$      & $1.75$                & \best{$3$}                 & \best{$15.63 \pm 0.20$} \\
            16000                         & 300                                         & $5.15$                                          & $76 \pm 1$                                  & $900 \pm 10$          & \best{$4.98$}         & $31$                       & $276 \pm 1$           & $5.14$                & \best{$6$}                 & \best{$22.68 \pm 0.08$} \\
            \bottomrule
        \end{tabular}}
    \caption{
        CPU benchmark results with mean $\pm$ standard deviation (shown only when non-zero) across 10 runs for a $25\times15$ rectangular map.
        \textbf{Bold} marks the best (lowest) value per metric and row.
    }
    \label{tab:benchmarks_cpu}
\end{table}

\begin{table}[tb]
    \centering
    \scriptsize
    \adjustbox{max width=\linewidth}{
        \begin{tabular}{@{}rr|ccc|ccc@{}}
            \toprule
            \multicolumn{2}{c|}{Data set} & \multicolumn{3}{c|}{\texttt{somoclu} (GPU)} & \multicolumn{3}{c}{\texttt{torchsom} (GPU)}                                                                                                                                       \\
            \cmidrule(lr){1-2} \cmidrule(lr){3-5} \cmidrule(lr){6-8}
            Samples                       & Features                                    & \gls{qe} $\downarrow$                       & \gls{te} (\%) $\downarrow$ & Time (s) $\downarrow$   & \gls{qe} $\downarrow$ & \gls{te} (\%) $\downarrow$ & Time (s) $\downarrow$   \\
            \midrule
            240                           & 4                                           & \best{$0.19$}                               & $23$                       & \best{$0.15 \pm 0.01$}  & $0.24$                & \best{$2$}                 & $1.29 \pm 0.04$         \\
            240                           & 50                                          & \best{$1.80$}                               & $57$                       & \best{$0.51 \pm 0.01$}  & $1.83$                & \best{$3$}                 & $1.37 \pm 0.04$         \\
            240                           & 300                                         & $5.34$                                      & $53$                       & $2.37 \pm 0.01$         & \best{$5.21$}         & \best{$14 \pm 2$}          & \best{$1.34 \pm 0.03$}  \\
            4000                          & 4                                           & \best{$0.16$}                               & $24$                       & \best{$1.77 \pm 0.02$}  & $0.23$                & \best{$1$}                 & $3.32 \pm 0.01$         \\
            4000                          & 50                                          & \best{$1.61$}                               & $37$                       & $8.05 \pm 0.05$         & $1.74$                & \best{$4$}                 & \best{$3.44 \pm 0.04$}  \\
            4000                          & 300                                         & \best{$5.02$}                               & $47$                       & $40.92 \pm 1.30$        & $5.13$                & \best{$7$}                 & \best{$3.33 \pm 0.03$}  \\
            16000                         & 4                                           & \best{$0.15$}                               & $22$                       & \best{$10.48 \pm 0.27$} & $0.23$                & \best{$0$}                 & $12.71 \pm 0.03$        \\
            16000                         & 50                                          & \best{$1.60$}                               & $30$                       & $31.82 \pm 0.16$        & $1.75$                & \best{$4$}                 & \best{$13.29 \pm 0.10$} \\
            16000                         & 300                                         & \best{$4.98$}                               & $37$                       & $161$                   & $5.14$                & \best{$5$}                 & \best{$13.08 \pm 0.04$} \\
            \bottomrule
        \end{tabular}}
    \caption{
        GPU benchmark results with mean $\pm$ standard deviation (shown only when non-zero) across 10 runs for a $25\times15$ rectangular map.
        \textbf{Bold} marks the best (lowest) value per metric and row.
    }
    \label{tab:benchmarks_gpu}
\end{table}

\texttt{torchsom}'s most consistent advantage is topology preservation: it attains the lowest \gls{te} in every rectangular configuration, on both CPU and GPU, reducing \gls{te} by 62\% to 100\% relative to the standard \texttt{MiniSom} baseline and remaining below \texttt{MiniSom}-JIT and \texttt{somoclu} throughout (e.g.\ at 16000\,$\times$\,300, \gls{te} of 6\% versus 76\%, 31\%, and 37\%). On hexagonal maps it leads in nearly all configurations, the only exceptions being the low-dimensional (4-feature) sets (Appendix~\ref{app:benchmark_results}).
\gls{qe} is comparable across backends, with \texttt{torchsom} being marginally higher.

On CPU (Table~\ref{tab:benchmarks_cpu}), \texttt{torchsom} is 81\% to 98\% faster than standard \texttt{MiniSom}: \texttt{MiniSom} updates one sample at a time and cannot exploit batched linear algebra, whereas \texttt{torchsom} processes the whole batch with vectorized tensor operations, so its advantage grows with both sample size and feature dimensionality.
Against the optimized \texttt{MiniSom}-JIT baseline the result is workload-dependent: \texttt{torchsom} matches or outperforms it in every rectangular configuration (up to $\sim$$16\times$), with the gap smallest on the low-dimensional (4-feature) sets, where there is little batched work to amortize.

            On GPU (Table~\ref{tab:benchmarks_gpu}), this batching effect combines with a startup overhead: \texttt{torchsom}'s reported GPU time carries a one-time initialization (independent of feature count), whereas on small problems \texttt{somoclu}'s compiled kernel starts up almost instantly.
            On the 4-feature and smallest sets there is too little compute to hide this overhead, so \texttt{somoclu} is faster; as the data grows it is amortized and \texttt{torchsom}'s parallelism dominates, reaching up to $\sim$$12\times$ faster than \texttt{somoclu} on high-dimensional data.
        The same fixed-cost-amortized-at-scale effect appears on CPU for \texttt{MiniSom}-JIT, whose one-time Numba compilation is only worthwhile once the workload is large enough to absorb it.
        Device choice is therefore itself workload-dependent: GPU execution is most beneficial for large, high-dimensional data (e.g.\ cutting the 16000\,$\times$\,300 time by roughly 40\% relative to CPU), while the multi-core CPU is competitive or faster for small or low-dimensional data.

        These advantages compound with map size.
        On the larger $90\times70$ grid (Appendix~\ref{app:benchmark_results}, roughly $17\times$ the neurons), \texttt{torchsom}'s CPU speedup over standard \texttt{MiniSom} grows to $\sim$$200\times$ and over \texttt{MiniSom}-JIT to $\sim$$50\times$,
        while its GPU speedup over \texttt{somoclu} reaches $\sim$$66\times$.
            \gls{te} stays markedly lower than every baseline, while \gls{qe} remains on par, as in the main comparison.
            The device trade-off also tips further toward the GPU: at this scale \texttt{torchsom}'s GPU run is $\sim$$8\times$ faster than its CPU run, confirming that the parallel backend is most valuable on large, high-dimensional problems.

        All reported \texttt{torchsom} times are conservative: the library additionally evaluates both metrics (\gls{qe} and \gls{te}) over the full batch at every epoch, an $\mathcal{O}(2 \times \text{batch} \times \text{epochs})$ cost that \texttt{MiniSom} and \texttt{somoclu} avoid by training only and measuring once at the end.
Overall, \texttt{torchsom} combines the strongest topology preservation with competitive or superior speed (dominant against standard \texttt{MiniSom}, and leading the optimized baselines on high-dimensional and GPU workloads) within a single modular library.

\section{Conclusion} \label{sec:conclusion}

We introduced \texttt{torchsom}, a \texttt{PyTorch}-native reference implementation of the \gls{som} that combines a \texttt{scikit-learn}-compatible API with native GPU acceleration, an advanced built-in visualization suite, a built-in clustering interface, and \gls{jitl} support within a single modular and openly licensed codebase.
Beyond the classical algorithm, \texttt{torchsom} provides a configurable \gls{bmu} search backend with optional \texttt{FAISS} compatibility, flexible neighborhood retrieval modes, and rectangular and hexagonal grid topologies that can be wrapped into toroidal structures through periodic boundary conditions.
Across nine synthetic benchmark configurations spanning sample sizes from 240 to 16{,}000 and feature dimensionalities from 4 to 300, and against fair per-device baselines, \texttt{torchsom} attains \gls{qe} parity with \texttt{MiniSom} and the lowest \gls{te} in every rectangular configuration (62\% to 100\% below standard \texttt{MiniSom}) and nearly all hexagonal ones.
It is 81\% to 98\% faster than standard \texttt{MiniSom} on CPU and, against the optimized \texttt{MiniSom}-JIT and \texttt{somoclu} baselines, is fastest on high-dimensional and GPU workloads while staying competitive elsewhere.
These advantages become more pronounced as the map size increases, with \texttt{torchsom} achieving speedups of up to $\sim200\times$ over \texttt{MiniSom} on CPU and $\sim66\times$ over \texttt{somoclu} on GPU (Appendix~\ref{app:benchmark_results}).

Several directions remain open for future work:
(i) integration of Growing and Hierarchical \gls{som} variants;
(ii) broader benchmarking, both against domain-specific data (e.g., time series, images, industrial sensor streams) beyond \texttt{scikit-learn}'s \texttt{make\_blobs()}, and against established dimensionality-reduction and manifold-learning baselines such as \texttt{PCA}, \texttt{t-SNE}, and \texttt{UMAP}; and
(iii) extension of the visualization suite with interactive widgets for exploratory analysis.

\texttt{torchsom} is distributed on \texttt{PyPI} under the \texttt{Apache 2.0} license, with narrative documentation at \url{https://opensource.michelin.io/TorchSOM} and source at \url{https://github.com/michelin/TorchSOM}.
The benchmark scripts and configuration files are released under the \texttt{benchmark/} directory of the repository for full reproducibility.
 
\clearpage
\bibliography{SOM_JMLR}

@misc{anselPyTorch2Faster2024,
  title = {{{PyTorch}} 2: {{Faster Machine Learning Through Dynamic Python Bytecode Transformation}} and {{Graph Compilation}}},
  shorttitle = {{{PyTorch}} 2},
  author = {Ansel, Jason and Yang, Edward and He, Horace and Gimelshein, Natalia and Jain, Animesh and Voznesensky, Michael and Bao, Bin and Bell, Peter and Berard, David and Burovski, Evgeni and Chauhan, Geeta and Chourdia, Anjali and Constable, Will and Desmaison, Alban and DeVito, Zachary and Ellison, Elias and Feng, Will and Gong, Jiong and Gschwind, Michael and Hirsh, Brian and Huang, Sherlock and Kalambarkar, Kshiteej and Kirsch, Laurent and Lazos, Michael and Lezcano, Mario and Liang, Yanbo and Liang, Jason and Lu, Yinghai and Luk, {\relax CK} and Maher, Bert and Pan, Yunjie and Puhrsch, Christian and Reso, Matthias and Saroufim, Mark and Siraichi, Marcos Yukio and Suk, Helen and Suo, Michael and Tillet, Phil and Wang, Eikan and Wang, Xiaodong and Wen, William and Zhang, Shunting and Zhao, Xu and Zhou, Keren and Zou, Richard and Mathews, Ajit and Chanan, Gregory and Wu, Peng and Chintala, Soumith},
  year = {2024},
  month = apr,
  journal = {29th ACM International Conference on Architectural Support for Programming Languages and Operating Systems, Volume 2 (ASPLOS '24)},
  doi = {10.1145/3620665.3640366},
  url = {https://docs.pytorch.org/assets/pytorch2-2.pdf},
  howpublished = {ACM}
}

@misc{berthierTorchSOMReferencePyTorch2025,
  title = {{{torchsom}}: {{The Reference PyTorch Library}} for {{Self-Organizing Maps}}},
  author = {Berthier, Louis},
  year = {2025},
  url = {https://github.com/michelin/TorchSOM},
  copyright = {Apache 2.0}
}

@article{bloomMARKETSEGMENTATIONNeural2005,
  title = {{{MARKET SEGMENTATION}}: {{A Neural Network Application}}},
  shorttitle = {{{MARKET SEGMENTATION}}},
  author = {Bloom, Jonathan Z.},
  year = {2005},
  month = jan,
  journal = {Annals of Tourism Research},
  volume = {32},
  number = {1},
  pages = {93--111},
  issn = {0160-7383},
  doi = {10.1016/j.annals.2004.05.001},
  url = {https://www.sciencedirect.com/science/article/pii/S0160738304001033}
}

@article{bowenSelforganizingMapsNovel2024,
  title = {Self-Organizing Maps: A Novel Approach to Identify and Map Business Clusters},
  shorttitle = {Self-Organizing Maps},
  author = {Bowen, Francis and Siegler, Jana{\'i}na},
  year = {2024},
  month = apr,
  journal = {Journal of Management Analytics},
  volume = {11},
  number = {2},
  pages = {228--246},
  publisher = {Taylor \& Francis},
  issn = {2327-0012},
  doi = {10.1080/23270012.2024.2306628},
  url = {https://doi.org/10.1080/23270012.2024.2306628}
}

@inproceedings{campelloDensityBasedClusteringBased2013,
  title = {Density-{{Based Clustering Based}} on {{Hierarchical Density Estimates}}},
  booktitle = {Advances in {{Knowledge Discovery}} and {{Data Mining}}},
  author = {Campello, Ricardo J. G. B. and Moulavi, Davoud and Sander, Joerg},
  year = {2013},
  pages = {160--172},
  publisher = {Springer},
  address = {Berlin, Heidelberg},
  doi = {10.1007/978-3-642-37456-2_14},
  isbn = {978-3-642-37456-2},
  langid = {english}
}

@misc{comitaniSimpSOMSimpleSelfOrganizing2017a,
  title = {{{SimpSOM}} ({{Simple Self-Organizing Maps}})},
  author = {Comitani, Federico},
  year = {2017},
  url = {https://github.com/fcomitani/simpsom},
  copyright = {GPL-3.0}
}

@article{concettiUnsupervisedAnomalyDetection2023,
  title = {An {{Unsupervised Anomaly Detection Based}} on {{Self-Organizing Map}} for the {{Oil}} and {{Gas Sector}}},
  author = {Concetti, Lorenzo and Mazzuto, Giovanni and Ciarapica, Filippo Emanuele and Bevilacqua, Maurizio},
  year = {2023},
  month = jan,
  journal = {Applied Sciences},
  volume = {13},
  number = {6},
  pages = {3725},
  publisher = {Multidisciplinary Digital Publishing Institute},
  issn = {2076-3417},
  doi = {10.3390/app13063725},
  url = {https://www.mdpi.com/2076-3417/13/6/3725},
  copyright = {http://creativecommons.org/licenses/by/3.0/},
  langid = {english}
}

@article{dashPerformanceAssessmentDifferent2024,
  title = {Performance {{Assessment}} of {{Different Sustainable Energy Systems Using Multiple-Criteria Decision-Making Model}} and {{Self-Organizing Maps}}},
  author = {Dash, Satyabrata and Chakravarty, Sujata and Giri, Nimay Chandra and Ghugar, Umashankar and Fotis, Georgios},
  year = {2024},
  month = mar,
  journal = {Technologies},
  volume = {12},
  number = {3},
  pages = {42},
  publisher = {Multidisciplinary Digital Publishing Institute},
  issn = {2227-7080},
  doi = {10.3390/technologies12030042},
  url = {https://www.mdpi.com/2227-7080/12/3/42},
  copyright = {http://creativecommons.org/licenses/by/3.0/},
  langid = {english}
}

@article{farzamniaMRIBrainTumor2023,
  title = {{{MRI Brain Tumor Detection Methods Using Contourlet Transform Based}} on {{Time Adaptive Self-Organizing Map}}},
  author = {Farzamnia, Ali and Hazaveh, Seyed Hamidreza and Siadat, Seyede Safieh and Moung, Ervin Gubin},
  year = {2023},
  journal = {IEEE Access},
  volume = {11},
  pages = {113480--113492},
  issn = {2169-3536},
  doi = {10.1109/ACCESS.2023.3322450},
  url = {https://ieeexplore.ieee.org/document/10273724/}
}

@article{fengAnalysisWaterQuality2023,
  title = {Analysis of Water Quality Indexes and Their Relationships with Vegetation Using Self-Organizing Map and Geographically and Temporally Weighted Regression},
  author = {Feng, Zhaohui and Xu, Chengjian and Zuo, Yiping and Luo, Xi and Wang, Lingqing and Chen, Hao and Xie, Xiaojing and Yan, Dan and Liang, Tao},
  year = {2023},
  month = jan,
  journal = {Environmental Research},
  volume = {216},
  pages = {114587},
  issn = {0013-9351},
  doi = {10.1016/j.envres.2022.114587},
  url = {https://www.sciencedirect.com/science/article/pii/S0013935122019144}
}

@article{figueiredoUnsupervisedLearningFinite2002,
  title = {Unsupervised Learning of Finite Mixture Models},
  author = {Figueiredo, M.A.T. and Jain, A.K.},
  year = {2002},
  month = mar,
  journal = {IEEE Transactions on Pattern Analysis and Machine Intelligence},
  volume = {24},
  number = {3},
  pages = {381--396},
  issn = {1939-3539},
  doi = {10.1109/34.990138},
  url = {https://ieeexplore.ieee.org/document/990138}
}

@article{gadJointSelfOrganizingMaps2024,
  title = {Joint {{Self-Organizing Maps}} and {{Knowledge-Distillation-Based Communication-Efficient Federated Learning}} for {{Resource-Constrained UAV-IoT Systems}}},
  author = {Gad, Gad and Farrag, Aya and Aboulfotouh, Ahmed and Bedda, Khaled and Fadlullah, Zubair Md. and Fouda, Mostafa M.},
  year = {2024},
  month = may,
  journal = {IEEE Internet of Things Journal},
  volume = {11},
  number = {9},
  pages = {15504--15522},
  issn = {2327-4662},
  doi = {10.1109/JIOT.2023.3349295},
  url = {https://ieeexplore.ieee.org/document/10379499}
}

@article{haoSOMDEScalableMethod2021a,
  title = {{{SOMDE}}: A Scalable Method for Identifying Spatially Variable Genes with Self-Organizing Map},
  shorttitle = {{{SOMDE}}},
  author = {Hao, Minsheng and Hua, Kui and Zhang, Xuegong},
  year = {2021},
  month = dec,
  journal = {Bioinformatics},
  volume = {37},
  number = {23},
  pages = {4392--4398},
  issn = {1367-4803},
  doi = {10.1093/bioinformatics/btab471},
  url = {https://doi.org/10.1093/bioinformatics/btab471}
}

@article{ikotunKmeansClusteringAlgorithms2023,
  title = {K-Means Clustering Algorithms: {{A}} Comprehensive Review, Variants Analysis, and Advances in the Era of Big Data},
  shorttitle = {K-Means Clustering Algorithms},
  author = {Ikotun, Abiodun M. and Ezugwu, Absalom E. and Abualigah, Laith and Abuhaija, Belal and Heming, Jia},
  year = {2023},
  month = apr,
  journal = {Information Sciences},
  volume = {622},
  pages = {178--210},
  issn = {0020-0255},
  doi = {10.1016/j.ins.2022.11.139},
  url = {https://www.sciencedirect.com/science/article/pii/S0020025522014633}
}

@article{jinEnsembleJustInTimeLearningBased2020,
  title = {Ensemble {{Just-In-Time Learning-Based Soft Sensor}} for {{Mooney Viscosity Prediction}} in an {{Industrial Rubber Mixing Process}}},
  author = {Jin, Huaiping and Li, Jiangang and Wang, Meng and Qian, Bin and Yang, Biao and Li, Zheng and Shi, Lixian},
  year = {2020},
  journal = {Advances in Polymer Technology},
  volume = {2020},
  number = {1},
  pages = {6575326},
  issn = {1098-2329},
  doi = {10.1155/2020/6575326},
  url = {https://onlinelibrary.wiley.com/doi/abs/10.1155/2020/6575326},
  copyright = {Copyright {\copyright} 2020 Huaiping Jin et al.},
  langid = {english}
}

@article{khanDiscoverBotnetsIoT2023,
  title = {Discover Botnets in {{IoT}} Sensor Networks: {{A}} Lightweight Deep Learning Framework with Hybrid Self-Organizing Maps},
  shorttitle = {Discover Botnets in {{IoT}} Sensor Networks},
  author = {Khan, Saad and Mailewa, Akalanka B.},
  year = {2023},
  month = mar,
  journal = {Microprocessors and Microsystems},
  volume = {97},
  pages = {104753},
  issn = {0141-9331},
  doi = {10.1016/j.micpro.2022.104753},
  url = {https://www.sciencedirect.com/science/article/pii/S0141933122002824}
}

@article{kohonenSelforganizedFormationTopologically1982a,
  title = {Self-Organized Formation of Topologically Correct Feature Maps},
  author = {Kohonen, Teuvo},
  year = {1982},
  month = jan,
  journal = {Biological Cybernetics},
  volume = {43},
  number = {1},
  pages = {59--69},
  issn = {1432-0770},
  doi = {10.1007/BF00337288},
  url = {https://doi.org/10.1007/BF00337288},
  langid = {english}
}

@article{kohonenSelforganizingMap1990,
  title = {The Self-Organizing Map},
  author = {Kohonen, T.},
  year = {1990},
  month = sep,
  journal = {Proceedings of the IEEE},
  volume = {78},
  number = {9},
  pages = {1464--1480},
  issn = {1558-2256},
  doi = {10.1109/5.58325},
  url = {https://ieeexplore.ieee.org/document/58325/}
}

@book{kohonenSelfOrganizingMaps2001a,
  title = {Self-{{Organizing Maps}}},
  author = {Kohonen, Teuvo},
  year = {2001},
  series = {Springer {{Series}} in {{Information Sciences}}},
  volume = {30},
  publisher = {Springer},
  address = {Berlin, Heidelberg},
  doi = {10.1007/978-3-642-56927-2},
  url = {http://link.springer.com/10.1007/978-3-642-56927-2},
  copyright = {http://www.springer.com/tdm},
  isbn = {978-3-540-67921-9 978-3-642-56927-2}
}

@article{licenSelforganizingMapAlgorithm2023,
  title = {Self-Organizing Map Algorithm for Assessing Spatial and Temporal Patterns of Pollutants in Environmental Compartments: {{A}} Review},
  shorttitle = {Self-Organizing Map Algorithm for Assessing Spatial and Temporal Patterns of Pollutants in Environmental Compartments},
  author = {Licen, Sabina and Astel, Aleksander and Tsakovski, Stefan},
  year = {2023},
  month = jun,
  journal = {Science of The Total Environment},
  volume = {878},
  pages = {163084},
  issn = {0048-9697},
  doi = {10.1016/j.scitotenv.2023.163084},
  url = {https://www.sciencedirect.com/science/article/pii/S0048969723017035}
}

@inproceedings{liMixtureDensityEstimation1999,
  title = {Mixture {{Density Estimation}}},
  booktitle = {Advances in {{Neural Information Processing Systems}}},
  author = {Li, Jonathan and Barron, Andrew},
  year = {1999},
  volume = {12},
  publisher = {MIT Press},
  url = {https://papers.nips.cc/paper_files/paper/1999/hash/a0f3601dc682036423013a5d965db9aa-Abstract.html}
}

@incollection{maltarolloApplicationsArtificialNeural2013,
  title = {Applications of {{Artificial Neural Networks}} in {{Chemical Problems}}},
  booktitle = {Artificial {{Neural Networks}} - {{Architectures}} and {{Applications}}},
  author = {Maltarollo, Vin{\'i}cius Gon{\c c}alves and Hon{\'o}rio, K{\'a}thia Maria and da Silva, Alb{\'e}rico Borges Ferreira and Maltarollo, Vin{\'i}cius Gon{\c c}alves and Hon{\'o}rio, K{\'a}thia Maria and da Silva, Alb{\'e}rico Borges Ferreira},
  year = {2013},
  month = jan,
  publisher = {IntechOpen},
  doi = {10.5772/51275},
  url = {https://www.intechopen.com/chapters/39067},
  isbn = {978-953-51-0935-8},
  langid = {english}
}

@article{mcinnesHdbscanHierarchicalDensity2017,
  title = {Hdbscan: {{Hierarchical}} Density Based Clustering},
  shorttitle = {Hdbscan},
  author = {McInnes, Leland and Healy, John and Astels, Steve},
  year = {2017},
  month = mar,
  journal = {Journal of Open Source Software},
  volume = {2},
  number = {11},
  pages = {205},
  issn = {2475-9066},
  doi = {10.21105/joss.00205},
  url = {https://joss.theoj.org/papers/10.21105/joss.00205},
  langid = {english}
}

@article{miaAnalysisSelforganizingMaps2023,
  title = {Analysis of Self-Organizing Maps and Explainable Artificial Intelligence to Identify Hydrochemical Factors That Drive Drinking Water Quality in {{Haor}} Region},
  author = {Mia, Md. Yousuf and Haque, Md. Emdadul and Islam, Abu Reza Md Towfiqul and Jannat, Jannatun Nahar and Jion, Most. Mastura Munia Farjana and Islam, Md. Saiful and Siddique, Md. Abu Bakar and Idris, Abubakr M. and Senapathi, Venkatramanan and Talukdar, Swapan and Rahman, Atiqur},
  year = {2023},
  month = dec,
  journal = {Science of The Total Environment},
  volume = {904},
  pages = {166927},
  issn = {0048-9697},
  doi = {10.1016/j.scitotenv.2023.166927},
  url = {https://www.sciencedirect.com/science/article/pii/S0048969723055523}
}

@misc{moosaviPythonLibrarySelf2014,
  title = {A {{Python Library}} for {{Self Organizing Map}} ({{SOM}})},
  author = {Moosavi, V and Packmann, S and Valles, I},
  year = {2014},
  url = {https://github.com/sevamoo/SOMPY},
  copyright = {Apache-2.0}
}

@misc{mullerSompbcSimpleSelforganizing2018,
  title = {Som-Pbc: {{A}} Simple Self-Organizing Map Implementation in {{Python}} with Periodic Boundary Conditions},
  author = {M{\"u}ller, Alex},
  year = {2018},
  url = {https://github.com/alexarnimueller/som}
}

@misc{pedregosaScikitlearnMachineLearning2011,
  title = {Scikit-Learn: {{Machine Learning}} in {{Python}}},
  shorttitle = {Scikit-Learn},
  author = {Pedregosa, Fabian and Varoquaux, Ga{\"e}l and Gramfort, Alexandre and Michel, Vincent and Thirion, Bertrand and Grisel, Olivier and Blondel, Mathieu and Prettenhofer, Peter and Weiss, Ron and Dubourg, Vincent and Vanderplas, Jake and Passos, Alexandre and Cournapeau, David and Brucher, Matthieu and Perrot, Matthieu and Duchesnay, {\'E}douard},
  year = {2011},
  journal = {Journal of Machine Learning Research},
  volume = {12},
  pages = {2825--2830},
  url = {https://jmlr.csail.mit.edu/papers/v12/pedregosa11a.html},
  copyright = {BSD-3-Clause}
}

@inproceedings{rajKeyGasesTransformer2023b,
  title = {Key {{Gases}} in {{Transformer Oil}} -- {{An Analysis}} Using {{Self Organizing Map}} ({{SOM}}) {{Neural Networks}}},
  booktitle = {2023 {{IEEE}} 12th {{International Conference}} on {{Communication Systems}} and {{Network Technologies}} ({{CSNT}})},
  author = {Raj, Raymon Antony and Sarathkumar, D and Andrews, Leo John Baptist and Venkatachary, Sampath Kumar},
  year = {2023},
  month = apr,
  pages = {642--647},
  issn = {2473-5655},
  doi = {10.1109/CSNT57126.2023.10134597},
  url = {https://ieeexplore.ieee.org/document/10134597}
}

@article{urhanIntegratingAdaptiveMoving2020,
  title = {Integrating Adaptive Moving Window and Just-in-Time Learning Paradigms for Soft-Sensor Design},
  author = {Urhan, Aysun and Alakent, Burak},
  year = {2020},
  month = jun,
  journal = {Neurocomputing},
  volume = {392},
  pages = {23--37},
  issn = {0925-2312},
  doi = {10.1016/j.neucom.2020.01.083},
  url = {https://www.sciencedirect.com/science/article/pii/S0925231220301417}
}

@misc{vettigliMiniSomMinimalisticNumPybased2018,
  title = {{{MiniSom}}: Minimalistic and {{NumPy-based}} Implementation of the {{Self Organizing Map}}},
  author = {Vettigli, Giuseppe},
  year = {2018},
  url = {https://github.com/JustGlowing/minisom},
  copyright = {MIT}
}

@article{weberApplicationSelforganizingMaps2023,
  title = {Application of Self-Organizing Maps to {{AFM-based}} Viscoelastic Characterization of Breast Cancer Cell Mechanics},
  author = {Weber, Andreas and dM Vivanco, Maria and {Toca-Herrera}, Jos{\'e} L.},
  year = {2023},
  month = feb,
  journal = {Scientific Reports},
  volume = {13},
  number = {1},
  pages = {3087},
  publisher = {Nature Publishing Group},
  issn = {2045-2322},
  doi = {10.1038/s41598-023-30156-3},
  url = {https://www.nature.com/articles/s41598-023-30156-3},
  copyright = {2023 The Author(s)},
  langid = {english}
}

@misc{wittekSomocluMassivelyParallel2017,
  title = {Somoclu: {{Massively}} Parallel Self-Organizing Maps},
  author = {Wittek, Peter},
  year = {2017},
  url = {https://github.com/peterwittek/somoclu},
  copyright = {MIT}
}

@article{xiangPotentialEcologicalRisk2022,
  title = {The Potential Ecological Risk Assessment of Soil Heavy Metals Using Self-Organizing Map},
  author = {Xiang, Qing and Yu, Huan and Chu, Hongliang and Hu, Mengke and Xu, Tao and Xu, Xiaoyu and He, Ziyi},
  year = {2022},
  month = oct,
  journal = {Science of The Total Environment},
  volume = {843},
  pages = {156978},
  issn = {0048-9697},
  doi = {10.1016/j.scitotenv.2022.156978},
  url = {https://www.sciencedirect.com/science/article/pii/S004896972204075X}
}

@article{zhangDeepSemiSupervisedJustinTime2022,
  title = {Deep {{Semi-Supervised Just-in-Time Learning Based Soft Sensor}} for {{Mooney Viscosity Estimation}} in {{Industrial Rubber Mixing Process}}},
  author = {Zhang, Yan and Jin, Huaiping and Liu, Haipeng and Yang, Biao and Dong, Shoulong},
  year = {2022},
  month = mar,
  journal = {Polymers},
  volume = {14},
  number = {5},
  pages = {1018},
  issn = {2073-4360},
  doi = {10.3390/polym14051018},
  url = {https://www.mdpi.com/2073-4360/14/5/1018},
  copyright = {https://creativecommons.org/licenses/by/4.0/},
  langid = {english}
}

@article{zhangHydrogeochemicalAnalysisGroundwater2023,
  title = {Hydrogeochemical Analysis and Groundwater Pollution Source Identification Based on Self-Organizing Map at a Contaminated Site},
  author = {Zhang, Yaobin and Zhang, Qiulan and Chen, Wenfang and Shi, Weiwei and Cui, Yali and Chen, Leilei and Shao, Jingli},
  year = {2023},
  month = jan,
  journal = {Journal of Hydrology},
  volume = {616},
  pages = {128839},
  issn = {0022-1694},
  doi = {10.1016/j.jhydrol.2022.128839},
  url = {https://www.sciencedirect.com/science/article/pii/S0022169422014093}
}

@inproceedings{zhangOnlineSemisupervisedQuality2018,
  title = {Online {{Semi-supervised Quality Prediction Model}} for {{Batch Mixing Process}}},
  booktitle = {2018 {{IEEE}} 7th {{Data Driven Control}} and {{Learning Systems Conference}} ({{DDCLS}})},
  author = {Zhang, Mingtao and Chen, Bocheng and Wu, You and Deng, Weiwei and Zhang, Xuelei and Liu, Yi},
  year = {2018},
  month = may,
  pages = {893--898},
  doi = {10.1109/DDCLS.2018.8516014},
  url = {https://ieeexplore.ieee.org/document/8516014}
}

@article{zhengJustintimeSemisupervisedSoft2018,
  title = {Just-in-Time Semi-Supervised Soft Sensor for Quality Prediction in Industrial Rubber Mixers},
  author = {Zheng, Wenjian and Liu, Yi and Gao, Zengliang and Yang, Jianguo},
  year = {2018},
  month = sep,
  journal = {Chemometrics and Intelligent Laboratory Systems},
  volume = {180},
  pages = {36--41},
  issn = {0169-7439},
  doi = {10.1016/j.chemolab.2018.07.002},
  url = {https://www.sciencedirect.com/science/article/pii/S0169743918300844}
}

@inproceedings{lamNumbaLLVMbasedPython2015,
  author = {Lam, Siu Kwan and Pitrou, Antoine and Seibert, Stanley},
  title = {Numba: a LLVM-based Python JIT compiler},
  year = {2015},
  isbn = {9781450340052},
  publisher = {Association for Computing Machinery},
  address = {New York, NY, USA},
  url = {https://doi.org/10.1145/2833157.2833162},
  doi = {10.1145/2833157.2833162},
  booktitle = {Proceedings of the Second Workshop on the LLVM Compiler Infrastructure in HPC},
  articleno = {7},
  numpages = {6},
  location = {Austin, Texas},
  series = {LLVM '15}
}

\clearpage
\appendix

\section{Glossary} \label{app:glossary}

\begin{description}
    \item[BMU] Best Matching Unit.
    \item[DL] Deep Learning.
    \item[JITL] Just-In-Time Learning.
    \item[ML] Machine Learning.
    \item[QE] Quantization Error.
    \item[SOM] Self-Organizing Map.
    \item[TE] Topographic Error.
\end{description}

\section{SOM Overview} \label{app:som_overview}

This appendix gives a self-contained, notation-complete description of the \gls{som} as implemented in \texttt{torchsom}.
The same material, kept in sync with the code, is mirrored in the online documentation at \url{https://opensource.michelin.io/TorchSOM/getting_started/basic_concepts.html}.

\textit{Setup and notation.}
A \gls{som} approximates a distribution over an input space $\mathcal{X} \subseteq \mathbb{R}^{d}$ by a two-dimensional lattice of $I \times J$ neurons.
Each neuron at grid position $(i,j)$, with $i \in \{1,\dots,I\}$ and $j \in \{1,\dots,J\}$, carries a \emph{codebook} (weight) vector $\mathbf{w}_{ij} \in \mathbb{R}^{l}$ with $l = d$, and the full parameter set is the tensor
\begin{equation}
    \label{eq:som_weights}
    \mathbf{W} \coloneqq [\mathbf{w}_{ij}]_{i \le I,\, j \le J} \in \mathbb{R}^{I \times J \times l}.
\end{equation}
Training uses a data set $\{\mathbf{x}_k\}_{k=1}^{N} \subset \mathbb{R}^{d}$ over epochs $t \in \{0,1,\dots,T\}$.

\textit{Best-matching unit and projection.}
Similarity in feature space is measured by a distance $\delta : \mathbb{R}^{d} \times \mathbb{R}^{l} \rightarrow \mathbb{R}_{\ge 0}$ (Eq.~\ref{eq:distances}).
For an input $\mathbf{x}$, the \gls{bmu} is the neuron whose codebook minimizes $\delta$,
\begin{equation}
    \label{eq:bmu}
    \mathrm{BMU}(\mathbf{x}) \coloneqq \operatorname*{arg\,min}_{(i,j)} \, \delta(\mathbf{x}, \mathbf{w}_{ij}),
\end{equation}
which induces a projection onto grid coordinates and a latent codebook retrieval,
\begin{equation}
    \label{eq:projection}
    \psi : \mathbb{R}^{d} \rightarrow \{1,\dots,I\} \times \{1,\dots,J\}, \qquad \psi(\mathbf{x}) \coloneqq \mathrm{BMU}(\mathbf{x}), \qquad \mathbf{z} \coloneqq \mathbf{w}_{\psi(\mathbf{x})} \in \mathbb{R}^{l}.
\end{equation}
The latent vector $\mathbf{z}$ is the representation used for clustering, visualization, and \gls{jitl} retrieval.

\textit{Competitive update.}
A \gls{som} learns by a neighborhood-weighted competitive rule rather than gradient descent: at each step the \gls{bmu} for the presented sample $\mathbf{x}$ is found, and each neuron is moved toward $\mathbf{x}$ by a step scaled by its grid proximity to the \gls{bmu},
\begin{equation}
    \label{eq:som_update}
    \mathbf{w}_{ij}(t+1) \coloneqq \mathbf{w}_{ij}(t) + \alpha(t)\, h_{ij}(t)\, \bigl(\mathbf{x} - \mathbf{w}_{ij}(t)\bigr),
\end{equation}
with learning rate $\alpha(t) \in \mathbb{R}_{\ge 0}$ (Eq.~\ref{eq:decay}) and neighborhood weight $h_{ij}(t) \in \mathbb{R}$ (Eq.~\ref{eq:neighborhood}).

\textit{Neighborhood function and order.}
Let $\rho \coloneqq d_{\mathrm{grid}}\bigl((i,j), \mathrm{BMU}\bigr)$ denote the \emph{grid-space} distance from neuron $(i,j)$ to the \gls{bmu}, where the grid metric is induced by the lattice geometry (Eq.~\ref{eq:pbc}).
\texttt{torchsom} provides four neighborhood kernels of width $\sigma(t)$,
\begin{equation}
    \label{eq:neighborhood}
    \begin{aligned}
        h^{\mathrm{gaussian}}_{ij}(t) & \coloneqq \exp\!\left(-\frac{\rho^{2}}{2\,\sigma(t)^{2}}\right),                                                   &
        h^{\mathrm{mexican}}_{ij}(t)  & \coloneqq \left(1 - \frac{\rho^{2}}{4\,\sigma(t)^{2}}\right)\exp\!\left(-\frac{\rho^{2}}{2\,\sigma(t)^{2}}\right),   \\
        h^{\mathrm{bubble}}_{ij}(t)   & \coloneqq \mathbb{I}\bigl(\rho \le \sigma(t)\bigr),                                                                &
        h^{\mathrm{triangle}}_{ij}(t) & \coloneqq \max\!\left(0,\; 1 - \frac{\rho}{\sigma(t)}\right).
    \end{aligned}
\end{equation}
Discrete neighborhoods are controlled by an integer \emph{order} $o \in \mathbb{N}^{+}$.
On a rectangular grid the order-$o$ neighborhood of the \gls{bmu} at $(i,j)$ is the Chebyshev ball
\begin{equation}
    \label{eq:neighborhood_order}
    \mathcal{N}_{o}\bigl((i,j)\bigr) \coloneqq \bigl\{ (i',j') \in \{1,\dots,I\} \times \{1,\dots,J\} : \max(|i'-i|,\, |j'-j|) \le o \bigr\},
\end{equation}
at most a $(2o{+}1)\times(2o{+}1)$ block of neurons; the hexagonal grid uses the analogous hop-distance rings (Figure~\ref{fig:som_topology}).
The order $o$ sets both the support of the discrete weight update and the sample-retrieval neighborhoods used for \gls{jitl} (the \texttt{neighborhood\_order} parameter and the \texttt{bmu\_neighborhood} retrieval modes).

\begin{figure}[bp]
    \centering
    \begin{subfigure}[b]{0.40\textwidth}
        \centering
        \begin{tikzpicture}[scale=0.5]
            \foreach \x in {1,...,7} {
                    \foreach \y in {1,...,7} {
                            \pgfmathtruncatemacro{\ord}{max(abs(\x-4),abs(\y-4))}
                            \ifcase\ord
                                \fillrect{\x}{\y}{bmucolor}\or
                                \fillrect{\x}{\y}{order1color}\or
                                \fillrect{\x}{\y}{order2color}\or
                                \fillrect{\x}{\y}{order3color}
                            \fi
                        }
                }
            \draw[step=1, gray!50, very thin] (1,1) grid (8,8);
        \end{tikzpicture}
        \caption{Rectangular: Chebyshev blocks.}
        \label{fig:som_topology_rect}
    \end{subfigure}
    \hfill
    \begin{subfigure}[b]{0.40\textwidth}
        \centering
        \begin{tikzpicture}
            \foreach \q in {-3,...,3} {
                    \foreach \r in {-3,...,3} {
                            \pgfmathtruncatemacro{\hd}{(abs(\q)+abs(\r)+abs(\q+\r))/2}
                            \ifnum\hd<4
                                \pgfmathsetmacro{\hx}{0.32*sqrt(3)*(\q+\r/2)}
                                \pgfmathsetmacro{\hy}{0.32*1.5*\r}
                                \ifcase\hd
                                    \node[regular polygon, regular polygon sides=6, fill=bmucolor, minimum size=0.62cm, inner sep=0pt, draw=white, line width=0.6pt] at (\hx,\hy) {}; \or
                                    \node[regular polygon, regular polygon sides=6, fill=order1color, minimum size=0.62cm, inner sep=0pt, draw=white, line width=0.6pt] at (\hx,\hy) {}; \or
                                    \node[regular polygon, regular polygon sides=6, fill=order2color, minimum size=0.62cm, inner sep=0pt, draw=white, line width=0.6pt] at (\hx,\hy) {}; \or
                                    \node[regular polygon, regular polygon sides=6, fill=order3color, minimum size=0.62cm, inner sep=0pt, draw=white, line width=0.6pt] at (\hx,\hy) {};
                                \fi
                            \fi
                        }
                }
        \end{tikzpicture}
        \caption{Hexagonal: hop-distance rings.}
        \label{fig:som_topology_hex}
    \end{subfigure}
    \caption{Neighborhood orders around the \gls{bmu} (black) for $o = 1$ (blue), $o = 2$ (teal), and $o = 3$ (pink), on the two grid topologies supported by \texttt{torchsom}, illustrating $\mathcal{N}_{o}$ in Eq.~\ref{eq:neighborhood_order}.}
    \label{fig:som_topology}
\end{figure}

\textit{Feature-space distances.}
The distance $\delta$ in Eq.~\ref{eq:bmu} is configurable; writing $x_a$ and $w_a$ for the $a$-th components,
\begin{equation}
    \label{eq:distances}
    \begin{aligned}
        \delta_{\mathrm{euclidean}}(\mathbf{x}, \mathbf{w}) & \coloneqq \sqrt{\textstyle\sum_{a=1}^{d} (x_a - w_a)^{2}},                                                 &
        \delta_{\mathrm{manhattan}}(\mathbf{x}, \mathbf{w}) & \coloneqq \textstyle\sum_{a=1}^{d} |x_a - w_a|,                                                              \\
        \delta_{\mathrm{cosine}}(\mathbf{x}, \mathbf{w})    & \coloneqq 1 - \frac{\mathbf{x} \cdot \mathbf{w}}{\lVert \mathbf{x} \rVert_2\, \lVert \mathbf{w} \rVert_2}, &
        \delta_{\mathrm{chebyshev}}(\mathbf{x}, \mathbf{w}) & \coloneqq \max_{1 \le a \le d} |x_a - w_a|.
    \end{aligned}
\end{equation}

\textit{Decay schedules.}
\texttt{torchsom} offers an inverse, a linear, and a general asymptotic schedule; one is selected per parameter, and each is a closed-form function of the epoch $t$ and of the initial value,
\begin{equation}
    \label{eq:decay}
    \begin{aligned}
        \alpha_{\mathrm{inv}}(t)  & \coloneqq \alpha(0)\,\frac{\gamma}{\gamma + t},     &
        \alpha_{\mathrm{lin}}(t)  & \coloneqq \alpha(0)\!\left(1 - \frac{t}{T}\right),        \\
        \sigma_{\mathrm{inv}}(t)  & \coloneqq \frac{\sigma(0)}{1 + t\,(\sigma(0)-1)/T}, &
        \sigma_{\mathrm{lin}}(t)  & \coloneqq \sigma(0) + t\,\frac{1 - \sigma(0)}{T},         \\
        \theta_{\mathrm{asym}}(t) & \coloneqq \frac{\theta(0)}{1 + 2t/T},               &   &
    \end{aligned}
\end{equation}
with total epochs $T \in \mathbb{N}$, current epoch $t \in \{0,\dots,T\}$, inverse-decay constant $\gamma \coloneqq T/100$, and $\theta \in \{\alpha,\sigma\}$ for the asymptotic schedule.
The linear schedules reach $\alpha_{\mathrm{lin}}(T) = 0$ and $\sigma_{\mathrm{lin}}(T) = 1$ exactly, as does $\sigma_{\mathrm{inv}}$.

\textit{Grid topologies and boundary conditions.}
\texttt{torchsom} supports rectangular and hexagonal grids.
On a hexagonal grid every neuron is equidistant from its six neighbors, whereas on a rectangular grid diagonal neighbors are farther away than orthogonal ones.
The lattice geometry fixes a planar embedding $\varphi$ of the grid positions: the identity on a rectangular grid, and the even-r offset layout on a hexagonal grid, in which odd rows are shifted by half a cell and consecutive rows are spaced by $\sqrt{3}/2$.
Either grid may enable periodic boundary conditions (the \texttt{pbc} flag), identifying opposite edges so the lattice becomes a torus.
Then, the grid metric is
\begin{equation}
    \label{eq:pbc}
    d_{\mathrm{grid}}\bigl((i,j),(i',j')\bigr) \coloneqq \min_{\mathbf{s} \in \mathcal{S}} \bigl\lVert \varphi(i,j) - \varphi(i',j') + \mathbf{s} \bigr\rVert_2,
\end{equation}
where $\mathcal{S}$ enumerates translations by the grid periods, so neighborhoods wrap across boundaries and corner neurons are not penalized.
Without PBC, $\mathcal{S} = \{\mathbf{0}\}$ and $d_{\mathrm{grid}}$ is the plain Euclidean distance in the embedding.

\textit{Initialization.}
Codebooks are initialized either
(i) at random, sampling each $\mathbf{w}_{ij}$ uniformly over the per-feature data range, or
(ii) by PCA, placing the $\mathbf{w}_{ij}$ on the plane spanned by the two leading principal components of $\{\mathbf{x}_k\}$.
PCA initialization speeds convergence and improves reproducibility.

\textit{Quality metrics.}
Map fidelity and topology preservation are quantified by the \gls{qe} and \gls{te} (reported in Tables~\ref{tab:benchmarks_cpu} and~\ref{tab:benchmarks_gpu}),
\begin{align}
    \mathrm{QE} & \coloneqq \frac{1}{N} \sum_{k=1}^{N} \bigl\lVert \mathbf{x}_k - \mathbf{w}_{\mathrm{BMU}(\mathbf{x}_k)} \bigr\rVert_2, \label{eq:quantization_error}                                  \\
    \mathrm{TE} & \coloneqq \frac{1}{N} \sum_{k=1}^{N} \mathbb{I}\!\left( \mathrm{BMU}_2(\mathbf{x}_k) \notin \mathcal{N}_1\bigl(\mathrm{BMU}(\mathbf{x}_k)\bigr) \right), \label{eq:topographic_error}
\end{align}
where $\mathcal{N}_1(\cdot)$ is the order-one neighborhood of the lattice (Eq.~\ref{eq:neighborhood_order} with $o = 1$): a neuron together with its immediate neighbors, the eight orthogonal and diagonal ones on a rectangular grid and the six at hop distance one on a hexagonal grid.
Every backend is scored with the same definition, so \gls{te} values are comparable within a table.
They are not comparable between the rectangular and hexagonal tables, whose adjacency sets differ in size.
\gls{qe} measures quantization fidelity, while \gls{te} measures the fraction of samples whose two best neurons are non-adjacent; lower is better for both.

The package structure, class hierarchy, \gls{bmu}-search backends, and end-to-end training flow are documented at \url{https://opensource.michelin.io/TorchSOM/user_guide/architecture.html}.

\section{\texttt{torchsom} Visualization Examples} \label{app:visualization}

\begin{figure}[hp]
    \centering
    \begin{subfigure}[t]{0.4\textwidth}
        \centering
        \includegraphics[width=\textwidth]{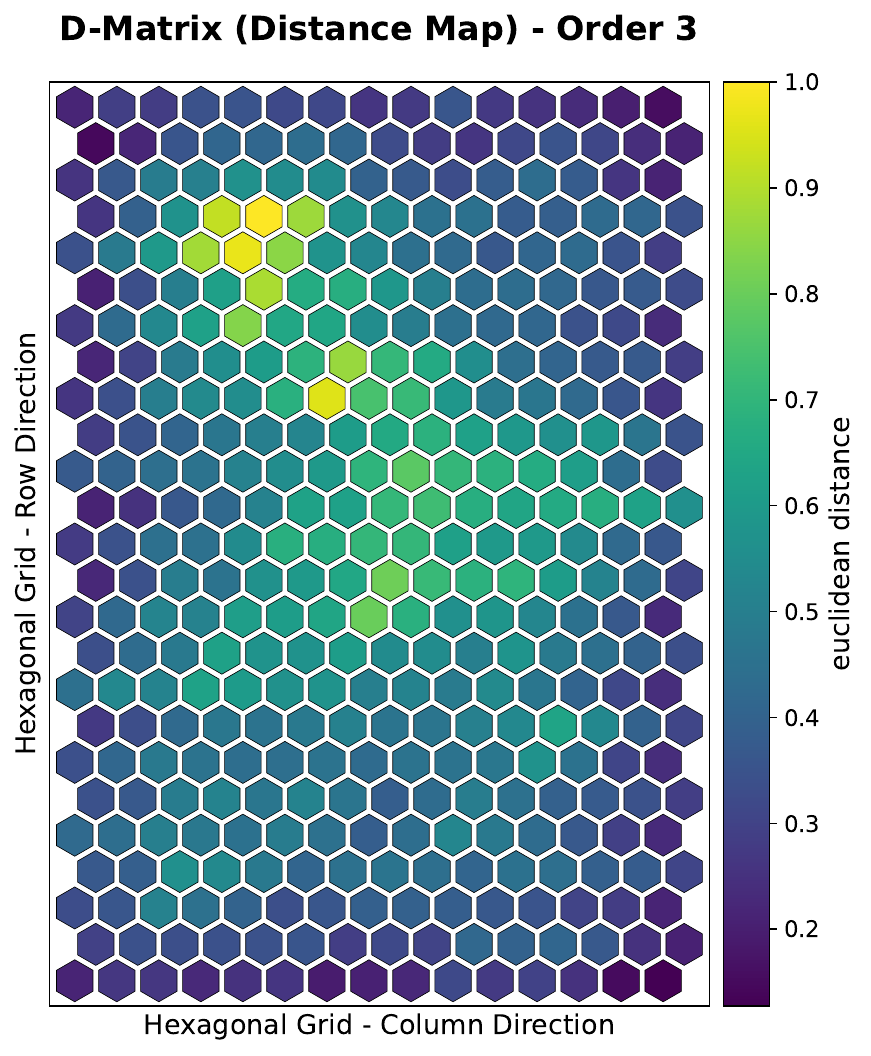}
        \caption{Distance map (U-matrix), wine data set.}
        \label{fig:appendix_c_distance_map}
    \end{subfigure}
    \hfill
    \begin{subfigure}[t]{0.4\textwidth}
        \centering
        \includegraphics[width=\textwidth]{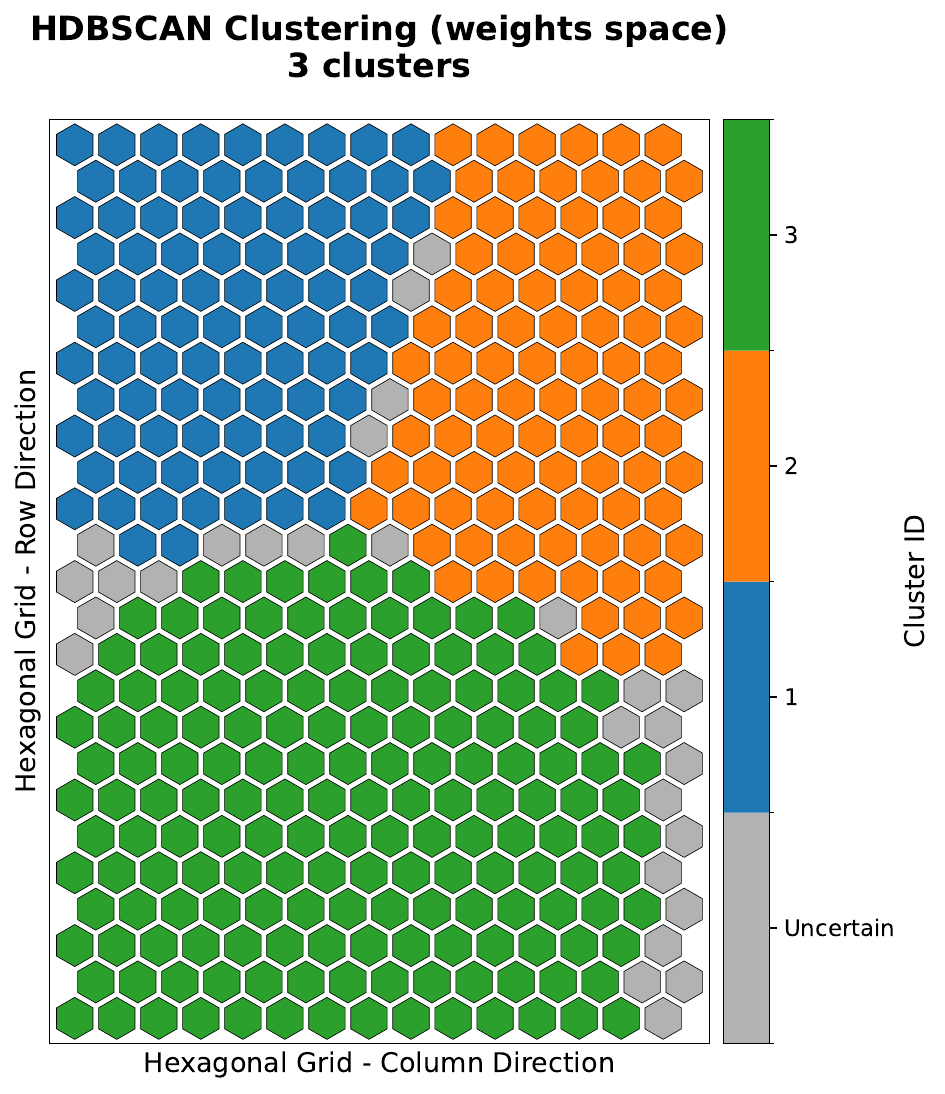}
        \caption{Cluster map, \texttt{make\_blobs} data set.}
        \label{fig:appendix_c_cluster_map}
    \end{subfigure}
    \caption[\texttt{torchsom} visualization examples]{
        Two outputs from \texttt{torchsom}'s integrated visualization suite (hexagonal topology):
        (a) a distance map (U-matrix) exposing cluster boundaries through inter-neuron distances; and
        (b) a cluster map showing the spatial assignment of automatically discovered clusters.
        The remaining visualizations are available at \url{https://opensource.michelin.io/TorchSOM/user_guide/visualization_help.html}.
    }
    \label{fig:appendix_c_examples}
\end{figure}

\texttt{torchsom} ships an integrated, \texttt{matplotlib}-based visualization suite covering seven categories of maps for both training diagnostics and post-hoc analysis (Table~\ref{tab:visualization_suite});
each category supports both rectangular and hexagonal topologies and integrates with the \texttt{scikit-learn}-style API.
Two representative outputs are shown in Figure~\ref{fig:appendix_c_examples}: a distance map (U-matrix) and a cluster map on two data sets.

\begin{table}[hp]
    \centering
    \small
    \adjustbox{max width=\linewidth}{%
        \begin{tabular}{@{}ll@{}}
            \toprule
            \textbf{Category}             & \textbf{Purpose}                                                                                   \\
            \midrule
            Learning curves               & Per-epoch \gls{qe}/\gls{te} traces during training                                                 \\
            Distance map (U-matrix)       & Inter-neuron distance landscape; reveals cluster boundaries                                        \\
            Hit map                       & Sample density / neuron utilization                                                                \\
            Component plane               & Per-feature weight surface across the grid                                                         \\
            Classification \& metric maps & Majority-class assignment (classification) and aggregated target statistics (regression)           \\
            Score \& rank maps            & Per-neuron reliability (variance, density, significance) and predicted-value ordering (regression) \\
            Cluster analysis              & Cluster map with silhouette, elbow, and clustering-algorithm comparison diagnostics                \\
            \bottomrule
        \end{tabular}}
    \caption{Visualization categories provided by \texttt{torchsom}'s built-in suite.}
    \label{tab:visualization_suite}
\end{table}

\section{Benchmark Results} \label{app:benchmark_results}

The tables below extend the main comparison (Tables~\ref{tab:benchmarks_cpu} and~\ref{tab:benchmarks_gpu}) to three additional configurations: a hexagonal $25\times15$ map across every sample/feature combination, and the larger $90\times70$ map at $300$ features in both rectangular and hexagonal topologies.
Each configuration follows the same device-matched layout as the main text: a CPU table with \texttt{MiniSom}, \texttt{MiniSom}-JIT, and \texttt{torchsom}, and a GPU table with \texttt{somoclu} and \texttt{torchsom}.

On both CPU and GPU the hexagonal results mirror the rectangular trends: \texttt{torchsom} again attains the lowest \gls{te} in nearly all configurations, the exception being the 4-feature sets, where \texttt{MiniSom}-JIT (CPU) and \texttt{somoclu} (GPU) are marginally lower.
\texttt{torchsom}'s hexagonal wall-clock time is markedly higher than on the rectangular map, which reflects a current implementation limitation rather than an algorithmic cost: the per-epoch \gls{te} evaluation is fully vectorized for rectangular grids but, for hexagonal grids, still iterates over the batch and copies each sample's best-matching units from GPU to CPU, and this host synchronization dominates the hexagonal runtime.
Therefore, the reported hexagonal times are conservative, and vectorizing this evaluation is a target for future optimization.

These extended tables confirm and sharpen the main-text findings at scale.
On the $90\times70$ map \texttt{torchsom}'s CPU speed advantage over standard \texttt{MiniSom} reaches $\sim$$200\times$, and $\sim$$66\times$ over \texttt{somoclu} on GPU, while \gls{te} stays lowest and \gls{qe} remains on par.
Rectangular and hexagonal maps agree, with one deviation: hexagonal GPU time is inflated by the unvectorized per-epoch \gls{te} evaluation noted above (e.g.\ $94.8$\,s versus $15.6$\,s rectangular at $16000\times300$), so those times remain conservative.

\begin{table}[p]
    \centering
    \small
    \adjustbox{max width=\linewidth}{
        \begin{tabular}{@{}rr|ccc|ccc|ccc@{}}
            \toprule
            \multicolumn{2}{c|}{Data set} & \multicolumn{3}{c|}{\texttt{MiniSom} (CPU)} & \multicolumn{3}{c|}{\texttt{MiniSom}-JIT (CPU)} & \multicolumn{3}{c}{\texttt{torchsom} (CPU)}                                                                                                                                                                                       \\
            \cmidrule(lr){1-2} \cmidrule(lr){3-5} \cmidrule(lr){6-8} \cmidrule(lr){9-11}
            Samples                       & Features                                    & \gls{qe} $\downarrow$                           & \gls{te} (\%) $\downarrow$                  & Time (s) $\downarrow$ & \gls{qe} $\downarrow$ & \gls{te} (\%) $\downarrow$ & Time (s) $\downarrow$   & \gls{qe} $\downarrow$ & \gls{te} (\%) $\downarrow$ & Time (s) $\downarrow$   \\
            \midrule
            240                           & 4                                           & \best{$0.18$}                                   & $25 \pm 6$                                  & $1.47 \pm 0.02$       & \best{$0.18$}         & \best{$22$}                & $1.98$                  & $0.24$                & $32$                       & \best{$0.50 \pm 0.03$}  \\
            240                           & 50                                          & \best{$1.78$}                                   & $56 \pm 5$                                  & $2.98 \pm 0.07$       & $1.79$                & $62$                       & $2.39 \pm 0.01$         & $1.80$                & \best{$15$}                & \best{$0.52 \pm 0.04$}  \\
            240                           & 300                                         & $5.39$                                          & $86 \pm 3$                                  & $12.95 \pm 0.03$      & $5.32$                & $53$                       & $5.72 \pm 0.02$         & \best{$5.22$}         & \best{$10$}                & \best{$0.72 \pm 0.04$}  \\
            4000                          & 4                                           & \best{$0.16$}                                   & $31 \pm 1$                                  & $23.47 \pm 0.38$      & \best{$0.16$}         & \best{$20$}                & \best{$5.64 \pm 0.04$}  & $0.23$                & $26$                       & $7.30 \pm 0.08$         \\
            4000                          & 50                                          & $1.66$                                          & $51 \pm 1$                                  & $45.86 \pm 2.13$      & \best{$1.64$}         & $27$                       & $14.58 \pm 0.76$        & $1.78$                & \best{$17$}                & \best{$7.64 \pm 0.11$}  \\
            4000                          & 300                                         & $5.12$                                          & $71 \pm 1$                                  & $233 \pm 1$           & \best{$5.02$}         & $34$                       & $68.86 \pm 0.51$        & $5.15$                & \best{$12 \pm 1$}          & \best{$12.65 \pm 0.17$} \\
            16000                         & 4                                           & \best{$0.16$}                                   & $31 \pm 1$                                  & $95.89 \pm 2.21$      & \best{$0.16$}         & \best{$19$}                & \best{$17.76 \pm 0.16$} & $0.22$                & $26$                       & $29.57 \pm 0.37$        \\
            16000                         & 50                                          & $1.64$                                          & $54 \pm 1$                                  & $184 \pm 2$           & \best{$1.61$}         & $24$                       & $51.02 \pm 0.22$        & $1.75$                & \best{$13$}                & \best{$30.03 \pm 0.18$} \\
            16000                         & 300                                         & $5.13$                                          & $72 \pm 1$                                  & $885 \pm 3$           & \best{$4.98$}         & $28$                       & $271 \pm 3$             & $5.15$                & \best{$11$}                & \best{$36.85 \pm 0.15$} \\
            \bottomrule
        \end{tabular}}
    \caption{
        CPU benchmark results with mean $\pm$ standard deviation (shown only when non-zero) across 10 runs for a $25\times15$ hexagonal map.
        \textbf{Bold} marks the best (lowest) value per metric and row.
    }
    \label{tab:benchmarks_small_hex_cpu}
\end{table}

\begin{table}[p]
    \centering
    \small
    \adjustbox{max width=\linewidth}{
        \begin{tabular}{@{}rr|ccc|ccc@{}}
            \toprule
            \multicolumn{2}{c|}{Data set} & \multicolumn{3}{c|}{\texttt{somoclu} (GPU)} & \multicolumn{3}{c}{\texttt{torchsom} (GPU)}                                                                                                                                       \\
            \cmidrule(lr){1-2} \cmidrule(lr){3-5} \cmidrule(lr){6-8}
            Samples                       & Features                                    & \gls{qe} $\downarrow$                       & \gls{te} (\%) $\downarrow$ & Time (s) $\downarrow$   & \gls{qe} $\downarrow$ & \gls{te} (\%) $\downarrow$ & Time (s) $\downarrow$   \\
            \midrule
            240                           & 4                                           & \best{$0.19$}                               & $43$                       & \best{$0.18$}           & $0.22$                & \best{$27$}                & $2.31 \pm 0.03$         \\
            240                           & 50                                          & \best{$1.78$}                               & $45$                       & \best{$0.54 \pm 0.02$}  & $1.84$                & \best{$18$}                & $2.89 \pm 0.03$         \\
            240                           & 300                                         & $5.29$                                      & $53$                       & \best{$2.46 \pm 0.02$}  & \best{$5.21$}         & \best{$12$}                & $2.91 \pm 0.04$         \\
            4000                          & 4                                           & \best{$0.17$}                               & \best{$26$}                & \best{$1.77 \pm 0.02$}  & $0.22$                & $29$                       & $20.65 \pm 0.04$        \\
            4000                          & 50                                          & \best{$1.62$}                               & $34$                       & \best{$7.71 \pm 0.19$}  & $1.75$                & \best{$14$}                & $20.56 \pm 0.11$        \\
            4000                          & 300                                         & \best{$5.01$}                               & $47$                       & $43.69 \pm 2.87$        & $5.15$                & \best{$13$}                & \best{$20.39 \pm 0.22$} \\
            16000                         & 4                                           & \best{$0.16$}                               & \best{$26 \pm 1$}          & \best{$8.34 \pm 0.07$}  & $0.22$                & $30$                       & $115$                   \\
            16000                         & 50                                          & \best{$1.61$}                               & $32 \pm 1$                 & \best{$33.23 \pm 0.13$} & $1.75$                & \best{$14$}                & $81.03 \pm 0.04$        \\
            16000                         & 300                                         & \best{$4.99$}                               & $38$                       & $155$                   & $5.15$                & \best{$13$}                & \best{$117 \pm 10$}     \\
            \bottomrule
        \end{tabular}}
    \caption{
        GPU benchmark results with mean $\pm$ standard deviation (shown only when non-zero) across 10 runs for a $25\times15$ hexagonal map.
        \textbf{Bold} marks the best (lowest) value per metric and row.
    }
    \label{tab:benchmarks_small_hex_gpu}
\end{table}

\begin{table}[p]
    \centering
    \small
    \adjustbox{max width=\linewidth}{
        \begin{tabular}{@{}rr|ccc|ccc|ccc@{}}
            \toprule
            \multicolumn{2}{c|}{Data set} & \multicolumn{3}{c|}{\texttt{MiniSom} (CPU)} & \multicolumn{3}{c|}{\texttt{MiniSom}-JIT (CPU)} & \multicolumn{3}{c}{\texttt{torchsom} (CPU)}                                                                                                                                                                                     \\
            \cmidrule(lr){1-2} \cmidrule(lr){3-5} \cmidrule(lr){6-8} \cmidrule(lr){9-11}
            Samples                       & Features                                    & \gls{qe} $\downarrow$                           & \gls{te} (\%) $\downarrow$                  & Time (s) $\downarrow$ & \gls{qe} $\downarrow$ & \gls{te} (\%) $\downarrow$ & Time (s) $\downarrow$ & \gls{qe} $\downarrow$ & \gls{te} (\%) $\downarrow$ & Time (s) $\downarrow$   \\
            \midrule
            240                           & 300                                         & $5.56$                                          & $78 \pm 6$                                  & $547 \pm 6$           & $5.51$                & $77$                       & $88.73 \pm 0.37$      & \best{$5.19$}         & \best{$11 \pm 1$}          & \best{$2.36 \pm 0.10$}  \\
            4000                          & 300                                         & $5.18$                                          & $84 \pm 1$                                  & $5526 \pm 572$        & \best{$5.09$}         & $67$                       & $1607 \pm 35$         & $5.11$                & \best{$7$}                 & \best{$36.56 \pm 0.17$} \\
            16000                         & 300                                         & $5.08$                                          & $85 \pm 1$                                  & $25860 \pm 613$       & \best{$4.96$}         & $62$                       & $6365 \pm 49$         & $5.12$                & \best{$7$}                 & \best{$130 \pm 2$}      \\
            \bottomrule
        \end{tabular}}
    \caption{
        CPU benchmark results with mean $\pm$ standard deviation (shown only when non-zero) across 10 runs for a $90\times70$ rectangular map.
        \textbf{Bold} marks the best (lowest) value per metric and row.
    }
    \label{tab:benchmarks_large_rect_cpu}
\end{table}

\begin{table}[p]
    \centering
    \small
    \adjustbox{max width=\linewidth}{
        \begin{tabular}{@{}rr|ccc|ccc@{}}
            \toprule
            \multicolumn{2}{c|}{Data set} & \multicolumn{3}{c|}{\texttt{somoclu} (GPU)} & \multicolumn{3}{c}{\texttt{torchsom} (GPU)}                                                                                                                                     \\
            \cmidrule(lr){1-2} \cmidrule(lr){3-5} \cmidrule(lr){6-8}
            Samples                       & Features                                    & \gls{qe} $\downarrow$                       & \gls{te} (\%) $\downarrow$ & Time (s) $\downarrow$ & \gls{qe} $\downarrow$ & \gls{te} (\%) $\downarrow$ & Time (s) $\downarrow$   \\
            \midrule
            240                           & 300                                         & $5.77$                                      & $82$                       & $38.46 \pm 4.71$      & \best{$5.20$}         & \best{$20$}                & \best{$3.20 \pm 0.04$}  \\
            4000                          & 300                                         & \best{$5.04$}                               & $60$                       & $471 \pm 10$          & $5.11$                & \best{$7$}                 & \best{$7.27 \pm 0.04$}  \\
            16000                         & 300                                         & \best{$4.95$}                               & $56$                       & $1038 \pm 15$         & $5.12$                & \best{$7$}                 & \best{$15.64 \pm 0.04$} \\
            \bottomrule
        \end{tabular}}
    \caption{
        GPU benchmark results with mean $\pm$ standard deviation (shown only when non-zero) across 10 runs for a $90\times70$ rectangular map.
        \textbf{Bold} marks the best (lowest) value per metric and row.
    }
    \label{tab:benchmarks_large_rect_gpu}
\end{table}

\begin{table}[bp]
    \centering
    \small
    \adjustbox{max width=\linewidth}{
        \begin{tabular}{@{}rr|ccc|ccc|ccc@{}}
            \toprule
            \multicolumn{2}{c|}{Data set} & \multicolumn{3}{c|}{\texttt{MiniSom} (CPU)} & \multicolumn{3}{c|}{\texttt{MiniSom}-JIT (CPU)} & \multicolumn{3}{c}{\texttt{torchsom} (CPU)}                                                                                                                                                                                     \\
            \cmidrule(lr){1-2} \cmidrule(lr){3-5} \cmidrule(lr){6-8} \cmidrule(lr){9-11}
            Samples                       & Features                                    & \gls{qe} $\downarrow$                           & \gls{te} (\%) $\downarrow$                  & Time (s) $\downarrow$ & \gls{qe} $\downarrow$ & \gls{te} (\%) $\downarrow$ & Time (s) $\downarrow$ & \gls{qe} $\downarrow$ & \gls{te} (\%) $\downarrow$ & Time (s) $\downarrow$   \\
            \midrule
            240                           & 300                                         & $5.56$                                          & $82 \pm 6$                                  & $315 \pm 2$           & $5.53$                & $70$                       & $87.30 \pm 0.30$      & \best{$5.21$}         & \best{$38$}                & \best{$2.62 \pm 0.06$}  \\
            4000                          & 300                                         & $5.14$                                          & $93 \pm 1$                                  & $5672 \pm 309$        & \best{$5.06$}         & $67$                       & $1426 \pm 16$         & $5.12$                & \best{$20$}                & \best{$48.93 \pm 1.44$} \\
            16000                         & 300                                         & $5.04$                                          & $83 \pm 1$                                  & $25250 \pm 2268$      & \best{$4.95$}         & $62$                       & $6410 \pm 107$        & $5.11$                & \best{$18$}                & \best{$148 \pm 6$}      \\
            \bottomrule
        \end{tabular}}
    \caption{
        CPU benchmark results with mean $\pm$ standard deviation (shown only when non-zero) across 10 runs for a $90\times70$ hexagonal map.
        \textbf{Bold} marks the best (lowest) value per metric and row.
    }
    \label{tab:benchmarks_large_hex_cpu}
\end{table}

\begin{table}[p]
    \centering
    \small
    \adjustbox{max width=\linewidth}{
        \begin{tabular}{@{}rr|ccc|ccc@{}}
            \toprule
            \multicolumn{2}{c|}{Data set} & \multicolumn{3}{c|}{\texttt{somoclu} (GPU)} & \multicolumn{3}{c}{\texttt{torchsom} (GPU)}                                                                                                                                     \\
            \cmidrule(lr){1-2} \cmidrule(lr){3-5} \cmidrule(lr){6-8}
            Samples                       & Features                                    & \gls{qe} $\downarrow$                       & \gls{te} (\%) $\downarrow$ & Time (s) $\downarrow$ & \gls{qe} $\downarrow$ & \gls{te} (\%) $\downarrow$ & Time (s) $\downarrow$   \\
            \midrule
            240                           & 300                                         & $5.69$                                      & $92$                       & $36.73 \pm 1.24$      & \best{$5.21$}         & \best{$35$}                & \best{$2.97 \pm 0.04$}  \\
            4000                          & 300                                         & \best{$5.01$}                               & $60$                       & $479 \pm 12$          & $5.11$                & \best{$19$}                & \best{$21.74 \pm 0.04$} \\
            16000                         & 300                                         & \best{$4.94$}                               & $55$                       & $1189 \pm 30$         & $5.12$                & \best{$19$}                & \best{$94.77 \pm 7.31$} \\
            \bottomrule
        \end{tabular}}
    \caption{
        GPU benchmark results with mean $\pm$ standard deviation (shown only when non-zero) across 10 runs for a $90\times70$ hexagonal map.
        \textbf{Bold} marks the best (lowest) value per metric and row.
    }
    \label{tab:benchmarks_large_hex_gpu}
\end{table}
 
\end{document}